\documentclass[11pt,a4paper]{article}
\usepackage[hyperref]{emnlp2020}
\usepackage{times}
\usepackage{latexsym}

\usepackage{microtype}

\usepackage{url}
\usepackage{amssymb}
\usepackage{enumitem} 
\usepackage{colortbl}
\usepackage{arydshln}
\usepackage{rotating}
\usepackage{pdflscape}
\usepackage{tabularx}
\usepackage{booktabs}
\usepackage{tcolorbox}
\usepackage{graphicx}
\usepackage{subcaption}
\usepackage{adjustbox}
\usepackage{placeins}
\usepackage{array,multirow,graphicx}
\tcbuselibrary{skins,breakable}

\definecolor{Gray}{gray}{0.85}
\newcolumntype{a}{>{\columncolor{Gray}}c}

\aclfinalcopy 

\usepackage[textsize=tiny]{todonotes}

\definecolor{LightCyan}{rgb}{0.88,1,1}

\newtcbox{\docspan}[1][]{
  enhanced, size=fbox, equal height group=spans,
  before=\adjustbox{valign=c}\bgroup,
  after=\egroup,#1
}

\title{The Eval4NLP Shared Task on Explainable Quality Estimation:\\ Overview and Results}

\author{Marina Fomicheva$^{*}$, 
Piyawat Lertvittayakumjorn$^{\dagger}$,
\\
\textbf{Wei Zhao}$^{\ddagger}$, 
\textbf{Steffen Eger}$^{\ddagger}$, 
\textbf{Yang Gao}$^{\diamond}$
\\
$*$ University of Sheffield, UK
\quad
$\dagger$ Imperial College London, UK
\\
$\ddagger$ TU Darmstadt, Germany
\quad
$\diamond$ Royal Holloway, University of London, UK
\\
\texttt{m.fomicheva@sheffield.ac.uk} \quad
\texttt{pl1515@imperial.ac.uk} \\
\texttt{wei.zhao@h-its.org} \quad
\texttt{eger@aiphes.tu-darmstadt.de}
\\
\texttt{yang.gao@rhul.ac.uk} 
}

\begin{document}
\maketitle
\begin{abstract}
In this paper, we introduce the Eval4NLP-2021 shared task on explainable quality estimation. Given a source-translation pair, this shared task requires not only to provide a sentence-level score indicating the overall quality of the translation, but also to \emph{explain} this score by identifying the words that negatively impact translation quality.
%
We present the data, annotation guidelines and evaluation setup of the shared task, describe
the six participating systems, and analyze the results.
%
To the best of our knowledge, this is the first shared task on explainable NLP evaluation metrics. 
Datasets and results are available at \url{https://github.com/eval4nlp/SharedTask2021}.
\end{abstract}

\section{Introduction}\label{sec:introduction}

\begin{figure*}
\centering
\begin{subfigure}[b]{0.4\textwidth}
\includegraphics[width=\textwidth]{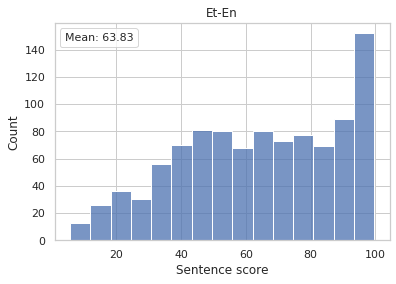}
\end{subfigure}
\begin{subfigure}[b]{0.4\textwidth}
\includegraphics[width=\textwidth]{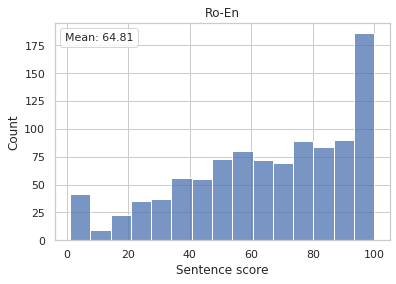}
\end{subfigure}
\begin{subfigure}[b]{0.4\textwidth}
\includegraphics[width=\textwidth]{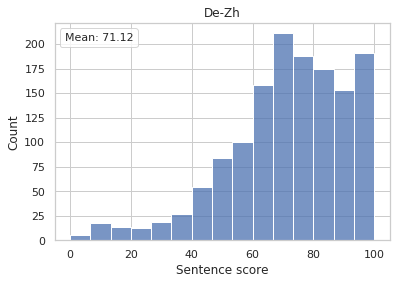}
\end{subfigure}
\begin{subfigure}[b]{0.4\textwidth}
\includegraphics[width=\textwidth]{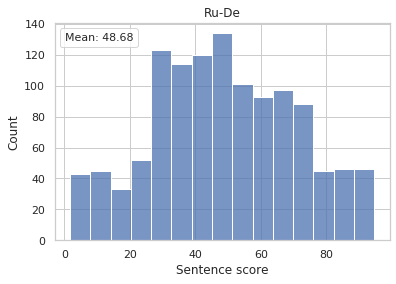}
\end{subfigure}
\caption{Distribution of sentence-level scores for each language pair.}
\label{fig:hist_DA}
\end{figure*}

Recent Natural Language Processing (NLP) systems based on pre-trained representations from Transformer language models, such as BERT \cite{devlin-etal-2019-bert} and XLM-Roberta \cite{conneau-etal-2020-unsupervised}, have achieved outstanding results in a variety of tasks. This boost in performance, however, comes at the cost of efficiency and interpretability. Interpretability is a major concern in modern Artificial Intelligence 
(AI) 
and NLP research \citep{doshi2017towards,danilevsky-etal-2020-survey}, as black-box models undermine users’ trust in new technologies \cite{mercado2016intelligent,toreini2020relationship}.

In the Eval4NLP 2021 shared task,
we focus on evaluating machine translation (MT) as an example of this problem. Specifically, we look at the task of \emph{quality estimation} (QE), where the aim is to predict the quality of MT output at inference time without access to reference translations \cite{blatz-etal-2004-confidence,specia2018quality}.\footnote{While QE is typically treated as a supervised task, a related research direction is \textit{reference-free} evaluation, which refers to unsupervised cross-lingual metrics that assess MT quality by computing distances between cross-lingual semantic representations of the source and target sentences \cite{zhao-etal-2020-limitations,song-etal-2021-sentsim}.}
Translation quality can be assessed at different levels of granularity: \emph{sentence-level}, i.e.\ predicting the overall quality of translated sentences, and \emph{word-level}, i.e.\ highlighting specific errors in the MT output. Those have traditionally been treated as two separate tasks, each one requiring dedicated training data.

In this shared task, we propose to address word-level translation error identification as an explainability task.\footnote{A  study on \emph{global explainability} of MT evaluation metrics,  disentangling them along linguistic factors such as syntax and semantics, has recently been conducted in \citet{kaster-et-al-2021-global}. In contrast, our shared task addresses \emph{local explainability} of individual input instances.} Explainability is a broad area aimed at explaining predictions of machine learning models. Rationale extraction methods achieve this by selecting a portion of the input that justifies model output for a given data point \cite{lei-etal-2016-rationalizing,jain-etal-2020-learning}. A natural way to explain sentence-level quality assessment is 
to identify translation errors. Hence, we frame \textbf{error identification as a task of providing explanations for the predictions of sentence-level QE models}. We claim that this task represents a challenging new benchmark for testing explainability for NLP and 
provides a new way of addressing word-level QE.

On the one hand, QE is different from other explainable NLP tasks with existing datasets \cite{deyoung2019eraser} in various important aspects. First, it is a regression task, as opposed to binary or multiclass text classification explored in previous work. Second, it is a multilingual task where the output score captures the relationship between source and target sentences. Finally, QE is fundamentally different from e.g.\ text classification, where clues are typically separate words or phrases \cite{zaidan2007using} that can often be considered independently of the rest of the text. By contrast, translation errors can only be identified given the context of the source and target sentences. Thus, this shared task provides a new benchmark for testing explainability methods in NLP.


On the other hand, treating word-level QE as an explainability problem offers some advantages compared to the current approaches. First, we can potentially avoid the need for supervised data at word level. Second, gold standard test sets can be made less expensive and more reliable. As we will show in Section \ref{sec:data}, rationalized sentence-level evaluation can be a middle ground between relatively cheap but noisy annotations derived from post-editing \cite{fomicheva2020mlqepe} and very informative but expensive explicit error annotation based on error taxonomies, such as the Multidimensional Quality Metrics (MQM) framework \cite{LommelMQM:2014}. For this shared task, we build a new test set with manually annotated explanations for sentence-level quality ratings. To the best of our knowledge, this is the first MT evaluation dataset annotated with human rationales.

The \textbf{main objective} of the shared task is threefold. 
First, it aims to explore the plausibility of explainable evaluation metrics \cite{wiegreffe-pinter-2019-attention}, by proposing a test set with manually annotated rationales. It helps the community better understand how similar the generated explanations are to the human explanations. 
Second, the shared task encourages research on unsupervised or semi-supervised methods for error identification, so as to reduce the cost on word-level MT error annotation.
Last but not least, the shared task sheds light on how current NLP evaluation systems arrive at their predictions and to what extent this process is aligned with human reasoning.

\section{Data}\label{sec:data}

For this shared task, we collected a new test set 
with (i) manual assessment of translation quality at sentence level and (ii) word-level rationales that explain the sentence-level scores (Section \ref{subsec:testset}). For training and development purposes, the participants were advised to use existing resources, which are briefly discussed in Section \ref{subsec:train_data}.

\subsection{Eval4NLP Test Set}
\label{subsec:testset}

\paragraph{Language pairs and MT systems}
The test set contains four language pairs: Estonian-English (Et-En), Romanian-English (Ro-En), Russian-German (Ru-De) and German-Chinese (De-Zh). For Et-En and Ro-En, we use the source and translated sentences from the test21 partition of the MLQE-PE dataset \cite{fomicheva2020mlqepe}. For Ru-De, the source sentences were extracted from Wikipedia following the procedure described in \citet{guzman-etal-2019-flores} and translated using the ML50 fairseq \cite{ott2019fairseq} multilingual Transformer model \cite{tang2020multilingual}. For De-Zh, the translations were produced using the Google Translate API, as the MT quality of the ML50 model was too low for this language pair according to our preliminary experiments.

\begin{table}[ht!]
\small
\begin{center}
\begin{tabular}{lcc}
\toprule
Language & Tokens & Sentences \\
pair & Source/Target & All/With rationales \\
\midrule
Et-En & 14,044/19,576 & 1,000/718 \\
Ro-En & 17,359/17,770 & 1,000/665 \\
De-Zh & 24,903/27,027 & 1,410/911 \\
Ru-De & 25,383/28,802 & 1,180/1,061 \\

\bottomrule
\end{tabular}
\caption{\label{tab:number_tokens} Total number of source tokens, target tokens, sentences and sentences with lower-than-perfect sentence score (i.e.\ sentences with rationales) in the Eval4NLP 2021 test set.}
\end{center}
\end{table}

\begin{figure*}
\centering
\includegraphics[width=0.8\textwidth]{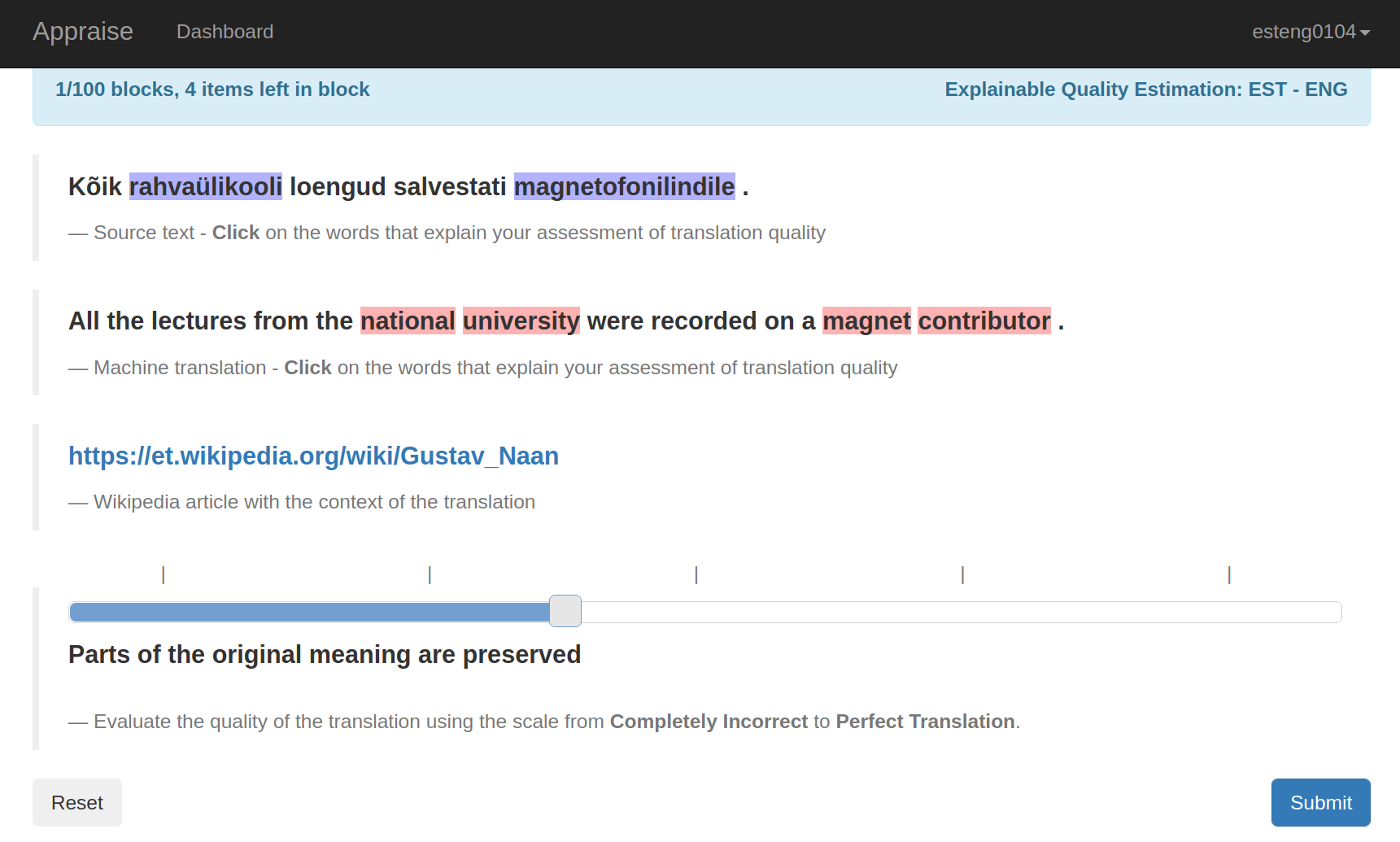}
\caption{Screenshot of the annotation interface.}
\label{fig:appraise}
\end{figure*}

\paragraph{Sentence- and word-level annotation}
For this annotation effort, we adapted the \emph{Appraise} manual evaluation interface \cite{mtm12_appraise}. For sentence-level annotation, we follow the guidelines from the MLQE-PE dataset \cite{fomicheva2020mlqepe}, a variant of the so called \emph{direct assessment} (DA) scores proposed by \citet{graham2016}. As illustrated in Figure \ref{fig:appraise}, the annotators were asked to provide a sentence rating by moving a slider on the quality scale from left (worse) to right (best). They were additionally provided with instructions on what specific quality ranges represent. Following \citet{graham2016}, the numeric values were not visible to the annotators, but the scale is interpreted numerically as follows: \textit{1-10} range represents a completely incorrect translation; \textit{11-30}, a translation that contains a few correct keywords, but the overall meaning is different or lost; \textit{31-50}, a translation that preserves parts of the original meaning; \textit{51-70}, a translation which is understandable and conveys the overall meaning of the source but contains a few errors; \textit{71-90}, a translation that closely preserves the semantics of the source and has only minor mistakes; and \textit{91-100}, a perfect translation. 

Crucially, besides the sentence-level rating, the annotators were asked to provide a rationale for their decisions. Specifically, for all translations except those they considered perfect, the annotators were required to highlight the words in the MT sentence corresponding to translation errors that would explain the assigned sentence score.\footnote{For all languages except for Chinese, the source sentences and MT outputs were tokenized with Moses tokenizer available at \url{https://github.com/moses-smt/mosesdecoder}. For Chinese, the jieba tokenizer was used: \url{https://github.com/fxsjy/jieba}.} They were also asked to highlight the source words that caused the errors in the MT output, as shown in Figure \ref{fig:appraise}. The missing contents was annotated by highlighting the source words that were not translated, whereas for the added (hallucinated) contents the annotators were only required to highlight the corresponding target words. We interpreted the highlighting as binary labels, indicating whether a given word is part of the rationale (positive class), or not (negative class). The annotators were provided with detailed annotation guidelines, which are available at \url{https://github.com/eval4nlp/SharedTask2021/tree/main/annotation-guidelines}. 

The annotation was conducted by 3 annotators for Et-En and Ro-En, and by (up to) 4 
annotators for Ru-De and De-Zh.\footnote{Not all annotators annotated all sentences for Ru-De and De-Zh. Individual annotators did 1411, 871, 1101, 1026 sentences for De-Zh, and 601, 1002, 1181, 1001 sentences for Ru-De.} Et-En and Ro-En data was annotated by Estonian and Romanian native speakers with near native proficiency in English. De-Zh data was annotated by Chinese native speakers with strong proficiency in German.
Finally, Ru-De data was annotated by native speakers of Russian with near native proficiency in German. The annotators for Ro-En, Et-En, and Ru-De are 
students specializing in Linguistics and Translation or are professional translators; the annotators for De-Zh are students specializing in computer science. 
The cost of annotation was approximately 4,000 Euro, with working times of 15 to 25 hours per annotator for De-Zh and Ru-De (Et-En and Ro-En annotators were compensated for the whole work, instead of on an hourly basis, and not all of them noted down their working times).

To produce a single sentence-level score, we take an average across the scores from individual annotators. To obtain a single binary label for each token, we use a majority voting mechanism, where the token is considered as part of the rationale if it was highlighted by the majority of the annotators.\footnote{When there is an even number of annotators, we weight the annotations by annotator reliability measured using their average agreement with the other annotators.}

\paragraph{Inter-annotator agreement} 
Table \ref{table:agreements} shows average agreement levels between our annotators (on common sets of annotated data instances). We use Pearson correlation for sentence-level scores 
and Cohen's kappa coefficient for word-level annotations. To be more precise, we measure Pearson correlation among all common instances between two annotators and then report the average across annotators; we measure average kappa agreement (averaged over all sentences) 
between any two annotators and then report the average across all annotators. We observe that Ro-En and Ru-De are most consistently annotated and De-Zh and Et-En have least agreement on average. Overall, our agreements are acceptable, however, in all cases, ranging from 0.42 to 0.67 kappa on word-level and $\sim$0.6 to 0.8 Pearson on sentence-level. For comparison, the average kappa reported by \citet{Lommel2014} for the fine-grained MQM error annotation ranges from 0.25 to 0.34.

\begin{table}[!htb]
    \small
    \centering
    \begin{tabular}{ccccc}
    \toprule
        & \textbf{Sentence-Level} & \multicolumn{2}{c}{\textbf{Word-Level}} \\
        & & Source & Target \\
    \midrule
    De-Zh & 0.59 & 0.48 & 0.55 \\
    Ru-De & 0.71 & 0.56 & 0.59 \\
    Ro-En & 0.81 & 0.54 & 0.67 \\
    Et-En & 0.68 & 0.42 & 0.45 \\
    \bottomrule
    \end{tabular}
    \caption{Sentence-level (Pearson) and word-level (kappa) agreements for different language pairs.}
    \label{table:agreements}
\end{table}

\paragraph{Data statistics}

\begin{table*}
    \small
    \centering
    \begin{tabular}{l rrr rrr r}
    \toprule
        Src & Pe 20 august , trupele sârbe au început urmărirea austriecilor în retragere . && &&& &\\
        PE & On 20 August , the Serbian troops began pursuing the retreating Austrians . && &&& &\\
    \bottomrule
    \end{tabular}

    
    \begin{tabular}{l l}
        MT & \colorbox{orange!0}{\strut Serbian} \colorbox{orange!0}{\strut troops} \colorbox{orange!0}{\strut started} \colorbox{orange!0}{\strut pursuing} \colorbox{orange!0}{\strut Austria} \colorbox{orange!0}{\strut on} \colorbox{orange!0}{\strut 20} \colorbox{orange!0}{\strut August} \colorbox{orange!0}{\strut in} \colorbox{orange!0}{\strut withdrawal} \colorbox{orange!0}{\strut .}\\
        
        Ann-EXPL & \colorbox{orange!0}{\strut Serbian} \colorbox{orange!0}{\strut troops} \colorbox{orange!0}{\strut started} \colorbox{orange!0}{\strut pursuing} \colorbox{orange!100}{\strut Austria} \colorbox{orange!0}{\strut on} \colorbox{orange!0}{\strut 20} \colorbox{orange!0}{\strut August} \colorbox{orange!100}{\strut in} \colorbox{orange!100}{\strut withdrawal} \colorbox{orange!0}{\strut .}\\
        
        Ann-EXPL* & \colorbox{orange!0}{\strut Serbian} \colorbox{orange!0}{\strut troops} \colorbox{orange!0}{\strut started} \colorbox{orange!0}{\strut pursuing} \colorbox{orange!66}{\strut Austria} \colorbox{orange!0}{\strut on} \colorbox{orange!33}{\strut 20} \colorbox{orange!33}{\strut August} \colorbox{orange!66}{\strut in} \colorbox{orange!100}{\strut withdrawal} \colorbox{orange!0}{\strut .}\\
        
        Predictions &  
        \colorbox{orange!5}{\strut Serbian} \colorbox{orange!0}{\strut troops} \colorbox{orange!2}{\strut started} \colorbox{orange!13}{\strut pursuing} \colorbox{orange!100}{\strut Austria} \colorbox{orange!1}{\strut on} \colorbox{orange!0}{\strut 20} \colorbox{orange!0}{\strut August} \colorbox{orange!36}{\strut in} \colorbox{orange!28}{\strut withdrawal} \colorbox{orange!5}{\strut .}\\ 
        
        Ann-PE & \colorbox{orange!100}{\strut Serbian} \colorbox{orange!100}{\strut troops} \colorbox{orange!100}{\strut started} \colorbox{orange!100}{\strut pursuing} \colorbox{orange!100}{\strut Austria} \colorbox{orange!100}{\strut on} \colorbox{orange!100}{\strut 20} \colorbox{orange!100}{\strut August} \colorbox{orange!100}{\strut in} \colorbox{orange!100}{\strut withdrawal} \colorbox{orange!0}{\strut .}\\\hline
        \\[-0.8em]
        
        Color scale& \colorbox{orange!0}{\strut0.0} \colorbox{orange!10}{\strut\quad} \colorbox{orange!20}{\strut\quad} \colorbox{orange!30}{\strut\quad} \colorbox{orange!40}{\strut\quad} \colorbox{orange!50}{\strut0.5} \colorbox{orange!60}{\strut\quad} \colorbox{orange!70}{\strut\quad} \colorbox{orange!80}{\strut\quad} \colorbox{orange!90}{\strut\quad} \colorbox{orange!100}{\strut1.0}\\
    \bottomrule
    \end{tabular}
    \caption{Example of the target-side annotation from the Ro-En test set and the output expected from the participants. ``Src" stands for the source sentence, ``MT" is the MT output, ``PE" is the post-edited version of the MT output taken from the MLQE-PE dataset. ``Ann-EXPL*" is the mean of the binary scores for each word averaged across the annotators. ``Ann-EXPL" corresponds to the binary scores obtained by aggregating individual annotations through majority voting (official gold standard of the shared task).  ``Ann-PE" is the word-level annotation derived from post-editing. ``Predictions" contains the predictions (after min-max normalization) for this sentence from the IST-Unbabel submission to the constrained track.}
    \label{tab:example}
\end{table*}

\begin{table}[t]
\small
\centering
\begin{tabular}{ll llll}
    \toprule
    & & Et-En & Ro-En & De-Zh & Ru-De \\
    \midrule
    \parbox[t]{2mm}{\multirow{2}{*}{\rotatebox[origin=c]{90}{Ours}}} & Source & 0.14 & 0.09 & 0.11 & 0.21 \\
    & MT & 0.18 & 0.13 & 0.12 & 0.21 \\
    \midrule
    \parbox[t]{2mm}{\multirow{3}{*}{\rotatebox[origin=c]{90}{MLQE}}} & & & & &  \\
    & Source & 0.22 & 0.20 & - & - \\
    & MT & 0.26 & 0.22 & - & - \\
    \bottomrule
\end{tabular}
\caption{Percentage of source and MT tokens annotated as rationales. For comparison, the percentage of source and target tokens annotated as errors in the same test partition of the MLQE-PE dataset is provided.}
\label{tab:perc_tags}
\end{table}

The number of annotated sentences, as well as the number of the source and target tokens in the test set are shown in Table \ref{tab:number_tokens}. In addition, we show the number of sentences with lower-than-perfect translation quality. This is the final subset of sentences that was used to evaluate the submissions to the shared task, since in our manual evaluation setup no rationales were required for the MT outputs with perfect quality. 
As shown in Table \ref{tab:number_tokens}, for all the language pairs the vast majority of translations has a lower-than-perfect score, where the percentage of such sentences is the lowest for De-Zh (65\%) and the highest for Ru-De (90\%).

Figure \ref{fig:hist_DA} shows the distribution of sentence-level scores for each language pair. The language pair with the highest average quality is De-Zh, whereas Ru-De has the lowest average score. For Et-En, Ro-En and Ru-De, the scores cover the whole quality range, while the distribution for De-Zh is highly skewed, which makes the task more challenging for this language pair (see Section \ref{sec:results}).

Table \ref{tab:perc_tags} shows the proportion of words annotated as rationales. The numbers in Table \ref{tab:perc_tags} are consistent with the average sentence-level quality, as De-Zh and Ru-De have the lowest and the highest percentage, respectively. This is expected given that lower quality translations should contain a higher number of errors. In general, the proportion of tokens considered relevant for explaining sentence-level ratings is fairly low. This is consistent with the annotation guidelines which stipulate that all and only the words necessary to justify the sentence score must be highlighted. Finally, we observe that, for Et-En and Ro-En, the percentage of annotated tokens is higher for the target than for the source sentences. This can be related to the presence of hallucinations, where the target contains words that do not have a clear correspondence with any part of the source sentence, as well as to typological differences between languages, whereby there tends to be a one-to-many correspondence between the source and target words.

\paragraph{Difference to existing QE datasets with word-level annotation}
The test set collected for this shared task is different from existing QE datasets with word-level annotation. A popular approach to building QE datasets is based on measuring post-editing effort \cite{bojar-EtAl:2017:WMT1,specia-etal-2018-findings,fonseca-etal-2019-findings,specia-etal-2020-findings-wmt}. This can be done at sentence level, by computing the so called HTER score \cite{snover2006study} that represents the minimum number of edits a human language expert is required to make in order to correct the MT output; or at word level, by aligning the MT output to its post-edited version and annotating the misaligned source and target words. An important limitation of this strategy is that the annotated words do not necessarily correspond to translation errors, as correcting a specific error may involve changing multiple related words in the sentence. This is exacerbated by the limitations of the heuristics used to automatically align the MT and its post-edited version. Indeed, as shown in Table \ref{tab:perc_tags}, the percentage of error tokens on the same data for Ro-En and Et-En language pairs is considerably higher in the MLQE-PE dataset, where word-level annotation is derived from post-editing.

An alternative approach is the explicit annotation of translation errors by human experts. This is typically done based on fine-grained error taxonomies such as the Multidimensional Quality Metrics (MQM) framework \cite{LommelMQM:2014}. While such annotations provide very informative labelled data, the annotator agreement for this style of annotation is fairly low \cite{Lommel2014} and the annotation is very time-consuming.\footnote{The interest towards MQM has recently increased due to a higher overall quality of MT \cite{freitag2021experts}, but the aforementioned issues still remain unsolved.}




The example in Table \ref{tab:example} shows a sample of the annotated data from the Ro-En test set. The first three rows correspond to the source (Src), the MT output (MT), and the post-edited MT output (PE). “Ann-EXPL*” shows the mean of the binary scores assigned by each annotator to a given word. Thus, the words “20” and “August” were included in the rationale by 1 out of 3 annotators for this example, the words “Austria” and “in” were highlighted by 2 out of 3 annotators; finally, the word “withdrawal” was included in the rationale by all 3 annotators. We can interpret this information as an indirect indication of error severity, as the most serious errors are expected to be noted by all of the annotators. “Ann-EXPL” shows the binary scores that we obtain through a majority voting  mechanism, as described above. These binary scores were used for the official evaluation reported in Section \ref{sec:results}. “Predictions” illustrates the predicted scores from one of the participants of the shared task.\footnote{We did not ask the participants to normalize the scores, as we are only interested in the ranking of tokens according to their relevance for sentence-level quality.} The predictions almost perfectly correspond to the human rationale, as in both cases the words “Austria” and “withdrawal” receive the highest scores. Finally, for comparison, “Ann-PE” shows the word labels for this sentence taken from the MLQE-PE dataset. In this case all tokens are considered as errors since re-orderings (or ``shifts") are not included in the set of possible edit operations used to compute minimum edit distance, from which the alignment between MT output and its PE is derived. 

To the best of our knowledge, this test set is the first MT evaluation dataset annotated with human rationales. The proposed annotation scheme has certain advantages for the QE task, as it allows to explicitly annotate translation errors, and at the same time results in higher agreement and less effort than fine-grained error annotation.\footnote{As shown by \citet{mcdonnell2017many},  rationales increase the reliability of human annotation when judging the relevance of webpages for information retrieval. In the future, we plan to investigate whether this also applies to MT evaluation and providing word-level explanations increases the consistency of sentence-level assessments.} 

\subsection{Training and development data}
\label{subsec:train_data}

\begin{table*}[ht!]
\small
\begin{center}
\begin{tabular}{r|p{9cm}|l}
\toprule
Team ID & Participating team \\ \midrule 
NICT-Kyoto & National Institute of Information and Communications Technology \hfill & \citet{rubino2021} \\
IST-Unbabel & IST/University of Lisbon \& Unbabel & \citet{treviso2021} \\
CLIP-UMD & Department of Computer Science, University of Maryland & \citet{kabir2021}\\
Gringham & Technical University of Darmstadt & \citet{leiter2021} \\
HeyTUDa & Technical University of Darmstadt & \citet{eksi2021} \\
CUNI-Prague & Charles University & \citet{polak2021} \\
\bottomrule
\end{tabular}
\end{center}
\caption{\label{tab:participants} Participants of the Eval4NLP Shared Task on Explainable Quality Estimation.}
\end{table*}

As discussed above, we use the same sentence-level annotation scheme as the one used in the MLQE-PE dataset. Therefore, for Ro-En and Et-En the participants could use the train and development partitions of MLQE-PE to build their sentence-level models. The De-Zh and Ru-De language pairs represent a fully zero-shot scenario where no sentence-level training data is available. 


\section{Task and Evaluation}\label{sec:task}
The task consisted of building a QE system that (i) predicts the quality score for an input pair of source text and MT hypothesis, (ii) provides word-level evidence for its predictions. An example of the test data used for evaluation is shown in Table \ref{tab:example}. 
The participants were expected to provide explanations for each sentence pair in the form of continuous scores, with the highest scores corresponding to the tokens considered as relevant by human annotators. The participants could submit to either \textbf{constrained} or \textbf{unconstrained} track. For the constrained track, the participants were expected to use no supervision at word level, while in the unconstrained track they were allowed to use any word-level data for training.

Explanations can be obtained either by building inherently interpretable models \cite{yu2019rethinking} or by using post-hoc explanation methods which extract explanations from an existing model \citep{Ribeiro:2016,Lundberg:2017,sundararajan2017axiomatic,schulz2020restricting}, for example by analysing the values of the gradient on each input feature. In this shared task, we provide both sentence-level training data and strong sentence-level models (see the TransQuest-LIME baseline in Section \ref{sec:baseline}), and thus encourage the participants to either train their own inherently interpretable models or use post-hoc techniques on top of our existing sentence-level models.

We accommodate the evaluation scheme to be suitable both for approaches that return continuous scores, and for supervised approaches that can return binary scores. Namely, we use evaluation metrics based on class probabilities that have been previously adapted for assessing the plausibility of rationale extraction methods \cite{atanasova2020diagnostic}. Since explainability methods typically proceed on instance-by-instance basis, and the scores produced for different instances are not necessarily comparable, we compute the evaluation metrics for each instance separately and average the results across all instances in the test set. Following \citet{fomicheva2021translation}, we define the following evaluation metrics to assess the performance of the submissions to the shared task at the word-level:
\paragraph{AUC score} For each instance, we compute the area under the receiver operating characteristic curve (AUC score) to compare the continuous attribution scores against binary gold labels. 
\paragraph{Average Precision} AUC scores can be overly optimistic for imbalanced data. Therefore, we also use Average Precision (AP). AP summarizes a precision-recall curve as the weighted mean of precisions achieved at each threshold, with the increase in recall from the previous threshold used as the weight \cite{Zhu2004}. 
\paragraph{Recall at Top-K} In addition, we report the Recall-at-Top-K metric commonly used in information retrieval. Applied to our setting, this metric computes the proportion of words with the highest attribution that correspond to translation errors against the total number of errors in the MT output. 
The code for computing the evaluation metrics can be found in the shared task github repository: \url{https://github.com/eval4nlp/SharedTask2021/tree/main/scripts}. The shared task used CodaLab as the submission platform.


\section{Baseline systems}\label{sec:baseline}

\paragraph{Random baseline} is built by sampling scores uniformly at random from a continuous $[0..1)$ range for each source and target token in a given sentence pair as well as for the sentence-level QE score.

\paragraph{Transquest-LIME} uses TransQuest QE models described in \citet{transquest:wmt20} to produce sentence-level scores. TransQuest follows the current standard practice of building task-specific NLP models by fine-tuning pre-trained multilingual language models, such as XLM-Roberta, on task-specific data. For Ro-En and Et-En, the Ro-En and Et-En TransQuest models are used, whereas for the zero-shot language pairs we use the multilingual variant of TransQuest, which was trained on a concatenation of MLQE-PE data. The post-hoc LIME explanation method \cite{Ribeiro:2016} is then applied to generate relevance scores for the source and target words. LIME is a simplification-based explanation technique, which fits a linear model in the vicinity of each test instance, to approximate the decision boundary of the complex model. Since in our sentence-level gold standard higher scores mean better quality, we invert LIME explanations so that higher values correspond to errors. 

\paragraph{XMover-SHAP} uses the reference-free metric XMoverScore \citep{zhao-etal-2020-limitations} to rate translations and uses the (likewise post-hoc) SHAP explainer  \cite{Lundberg:2017} to explain the ratings. In particular, given a source-translation pair, XMoverScore provides a real number to indicate the quality of the translation, in terms of its semantic overlapping with the source sentence, using re-mapped multilingual BERT embeddings and a target-side language model.\footnote{Note that XMoverScore is an unsupervised reference-free metric, in contrast to the supervised TransQuest QE model.} 
To explain the contribution of each
word in the rating, SHAP creates perturbations
of the source/translation sentence by
masking out some words and estimates 
the average marginal contribution of each word across 
all possible perturbations.
The source code for all the baseline systems 
is available at \url{https://github.com/eval4nlp/SharedTask2021/tree/main/baselines}.

\section{Participants}\label{sec:participants}

For this first edition of the shared task, we had a total of 6 participating teams listed in Table \ref{tab:participants}.\footnote{Initially, there were seven participating teams, but one of them opted out after the competition ended.} Below, we briefly describe the submitted approaches.

\paragraph{NICT-Kyoto}
use synthetic data to fine-tune the XLM-Roberta language model for the QE task. To produce synthetic sentence-level scores, they translate publicly available parallel corpora using SOTA neural MT systems and compute three reference-based metrics: ChrF \cite{popovic:2015:WMT2015}, TER \cite{snover2006study} and BLEU \cite{papineni2002bleu}. To simulate word-level annotation, they derive word-level labels from the alignment between the MT outputs and human reference translations. The QE model is then jointly trained to predict the scores from different metrics as well as word-level tags. A metric embedding component is proposed where each metric is represented with a set of learnable parameters. An attention mechanism between the metric embeddings and the input representations is employed to obtain word-level scores as explanations for the sentence-level predictions.

\paragraph{IST-Unbabel}
participated in the constrained and unconstrained tracks of the shared task. For the constrained track ("IST-Unbabel" in Table \ref{tab:results}), they used a set of explainability methods to extract the relevance of the input tokens from sentence-level QE models built on top of XLM-Roberta and RemBERT. The explainability methods explored in this work include attention-based, gradient-based and perturbation based approaches, as well as rationalization by construction. The best performing method which was submitted to the competition relies on the attention mechanism of the pre-trained Transformers in order to obtain the relevance scores for each token. In addition, scaling attention weights by the L2 norm of value vectors as suggested in \citet{kobayashi2020attention} resulted in a further boost in performance. 

For the unconstrained track ("IST-Unbabel*" in Table \ref{tab:results}), they add a word-level loss to the sentence-level models and train jointly using the annotated data from the MLQE-PE dataset.

\paragraph{HeyTUDa}
use the TransQuest QE models \cite{transquest:wmt20} for sentence-level prediction and a set of explainability techniques to estimate the relevance of each source and target word. Specifically, they explore three perturbation-based methods:
LIME, 
SHAP, 
and occlusion \cite{Zeiler:2014}, as well as three gradient-based methods: DeepLift \citep{deeplift}, Layer Gradient x Activation \citep{Shrikumar2016} and Integrated Gradients \citep{Sundararajan2017}. They further use an unsupervised ensembling method to combine the different explainability approaches. 

\paragraph{Gringham}
use the reference-free metrics XBERTScore (i.e., BERTScore \citep{Zhang2020BERTScore:2020} with cross-lingual embeddings) and XMoverScore and make them inherently interpretable by considering the token alignments produced by the models. The intuition is that words that are not well-aligned are most likely erroneous. Specifically, they explore XBERTScore and XMoverScore as sentence-level models and use the corresponding similarity (or distance) matrices to produce token-level scores.  

\paragraph{CLIP-UMD}
propose an ensemble of two approaches: (1) the LIME explanation technique applied to the TransQuest sentence-level model; (2) Divergent mBERT \cite{briakou-carpuat-2020-detecting}, which is a BERT-based model that can detect cross-lingual semantic divergences. Divergent mBERT is trained using synthetic data where semantic divergences are introduced automatically following a set of pre-defined perturbations. To produce a combination of the two methods, the predictions from each approach are averaged. 

\paragraph{CUNI-Prague}
participated in the unconstrained track. They fine-tune the XLM-R model for word-level and sentence-level QE. To map sentence piece tokenization from XLM-R to Moses tokenization, they ignore all sentence piece tokens corresponding to a given Moses token except the first one.

\section{Results}\label{sec:results}

\begin{table*}[ht!]
\small
\centering
\begin{tabular}{lacccacccc}
    \rowcolor{white}
    \toprule
	& \multicolumn{4}{c}{Target} & \multicolumn{4}{c}{Source} & \multicolumn{1}{c}{Sentence}  \\
	& Rank & AUC & AP & Rec & Rank & AUC & AP & Rec & Pearson \\
	\midrule
	& \multicolumn{9}{c}{Estonian-English} \\
	\midrule
	IST-Unbabel* & 1.0 & \textbf{0.92} & \textbf{0.85} & \textbf{0.76} & 1.3 & \textbf{0.94} & \textbf{0.86} & \textbf{0.77} & 0.86\\
    CUNI Prague* & 2.0 & 0.92 & 0.84 & \textbf{0.75} & 3.0 & 0.93 & \textbf{0.85} & \textbf{0.76} & 0.80\\
    NICT Kyoto$^\dagger$ & 3.0 & \textbf{0.90} & 0.82 & \textbf{0.73} & 1.7 & 0.93 & 0.85 & \textbf{0.77} & 0.85\\
    IST-Unbabel & 4.3 & 0.82 & 0.74 & 0.63 & 4.0 & 0.86 & 0.76 & 0.64 & 0.82\\
    Gringham & 4.7 & 0.84 & 0.71 & 0.60 & 5.0 & 0.86 & 0.72 & 0.59 & 0.71\\
    CLIP-UMD & 6.0 & 0.74 & 0.63 & 0.53 & 6.0 & 0.76 & 0.51 & 0.45 & 0.77\\
    Baseline: TransQuest-LIME & 7.3 & 0.62 & 0.54 & 0.43 & 7.0 & 0.54 & 0.44 & 0.31 & 0.77\\
    HeyTUDa & 7.7 & 0.66 & 0.52 & 0.41 & 10. & N/A & N/A & N/A & 0.77\\
    Baseline: XMover-SHAP & 9.0 & 0.62 & 0.44 & 0.34 & 8.0 & 0.54 & 0.37 & 0.23 & 0.49\\
    Baseline: Random & 10. & 0.50 & 0.36 & 0.25 & 9.0 & 0.49 & 0.34 & 0.19 & -0.03\\
    \midrule
    & \multicolumn{9}{c}{Romanian-English} \\
    \midrule
    NICT Kyoto$^\dagger$ & 1.0 & \textbf{0.95} & \textbf{0.87} & \textbf{0.78} & 1.0 & \textbf{0.95} & \textbf{0.85} & \textbf{0.75} & 0.92\\
    IST-Unbabel* & 2.0 & 0.94 & 0.84 & 0.75 & 2.0 & 0.93 & 0.81 & 0.71 & 0.87\\
    CUNI Prague* & 3.0 & 0.94 & 0.83 & 0.73 & 3.0 & 0.93 & 0.81 & 0.70 & 0.89\\
    IST-Unbabel & 4.0 & 0.88 & 0.78 & 0.68 & 4.0 & 0.86 & 0.73 & 0.62 & 0.90\\
    Gringham & 5.0 & 0.87 & 0.73 & 0.61 & 5.0 & 0.84 & 0.61 & 0.45 & 0.78\\
    CLIP-UMD & 6.0 & 0.73 & 0.60 & 0.49 & 6.0 & 0.72 & 0.41 & 0.37 & 0.90\\
    Baseline: TransQuest-LIME & 7.7 & 0.63 & 0.52 & 0.42 & 7.7 & 0.48 & 0.35 & 0.24 & 0.90\\
    HeyTUDa & 7.7 & 0.68 & 0.50 & 0.38 & 10. & N/A & N/A & N/A & 0.90\\
    Baseline: XMover-SHAP & 8.7 & 0.67 & 0.44 & 0.30 & 8.0 & 0.53 & 0.29 & 0.15 & 0.70\\
    Baseline: Random & 10. & 0.52 & 0.31 & 0.19 & 8.3 & 0.50 & 0.28 & 0.15 & 0.02\\
    \midrule
    & \multicolumn{9}{c}{Russian-German} \\
    \midrule
    NICT Kyoto$^\dagger$ & 1.0 & \textbf{0.93} & \textbf{0.83} & \textbf{0.74} & 1.0 & \textbf{0.92} & \textbf{0.80} & \textbf{0.71} & 0.68\\
    IST-Unbabel* & 2.0 & 0.80 & 0.64 & 0.52 & 2.0 & 0.85 & 0.71 & 0.59 & 0.67\\
    CUNI Prague* & 3.3 & 0.76 & 0.61 & 0.50 & 3.3 & 0.80 & 0.67 & 0.56 & 0.61\\
    Gringham & 4.3 & 0.79 & 0.57 & 0.46 & 3.7 & 0.84 & 0.67 & 0.56 & 0.60\\
    IST-Unbabel & 4.3 & 0.75 & 0.58 & 0.47 & 5.0 & 0.77 & 0.63 & 0.52 & 0.64\\
    CLIP-UMD & 6.0 & 0.65 & 0.46 & 0.36 & 6.3 & 0.66 & 0.41 & 0.37 & 0.30\\
    HeyTUDa & 7.3 & 0.54 & 0.33 & 0.23 & 10. & N/A & N/A & N/A & 0.50\\
    Baseline: XMover-SHAP & 7.7 & 0.52 & 0.33 & 0.23 & 8.0 & 0.52 & 0.36 & 0.26 & 0.25\\
    Baseline: Random & 9.0 & 0.49 & 0.31 & 0.22 & 9.0 & 0.51 & 0.34 & 0.24 & -0.02\\
    Baseline: TransQuest-LIME & 10. & 0.40 & 0.26 & 0.16 & 6.7 & 0.53 & 0.43 & 0.32 & 0.50\\
    \midrule
    & \multicolumn{9}{c}{German-Chinese} \\
    \midrule
    NICT Kyoto$^\dagger$ & 1.0 & \textbf{0.85} & \textbf{0.68} & \textbf{0.57} & 1.0 & \textbf{0.85} & \textbf{0.64} & \textbf{0.51} & 0.29\\
    IST-Unbabel & 2.3 & 0.68 & 0.50 & 0.37 & 3.0 & 0.67 & 0.47 & 0.32 & 0.33\\
    IST-Unbabel* & 2.7 & 0.71 & 0.47 & 0.34 & 2.0 & 0.73 & 0.50 & 0.35 & 0.27\\
    CUNI Prague* & 4.3 & 0.61 & 0.44 & 0.30 & 5.0 & 0.62 & 0.42 & 0.27 & 0.25\\
    CLIP-UMD & 4.7 & 0.63 & 0.40 & 0.27 & 6.3 & 0.61 & 0.31 & 0.25 & 0.50\\
    Gringham & 6.3 & 0.55 & 0.36 & 0.22 & 4.0 & 0.64 & 0.42 & 0.27 & -0.04\\
    Baseline: XMover-SHAP & 6.7 & 0.55 & 0.33 & 0.22 & 9.0 & 0.47 & 0.29 & 0.16 & 0.18\\
    HeyTUDa & 8.0 & 0.51 & 0.31 & 0.18 & 10. & N/A & N/A & N/A & 0.34\\
    Baseline: Random & 9.0 & 0.50 & 0.29 & 0.17 & 7.7 & 0.50 & 0.30 & 0.17 & 0.00\\
    Baseline: TransQuest-LIME & 10. & 0.46 & 0.27 & 0.14 & 7.0 & 0.49 & 0.32 & 0.20 & 0.34\\
    \bottomrule
\end{tabular}
\caption{Official results of the Eval4NLP Shared Task on Explainable Quality Estimation. Submissions to the unconstrained track are marked with *. We mark the NICT Kyoto submissions with a $^\dagger$, as they submitted to the constrained track, but use synthetic data for word-level supervision. 
Submissions not significantly outperformed by any other submission according to paired t-test for each metric are marked in bold.
N/A means that the participating team did not submit the word-level scores for the source sentences.}
\label{tab:results}
\end{table*}

Table \ref{tab:results} shows 
the results of the shared task. We report the word-level metrics presented in Section \ref{sec:task}, as well as Pearson correlation at sentence level.
The values of the ``Rank'' columns are computed by first ranking the participants according to each of the three word-level metrics and then averaging the resulting rankings.\footnote{This ranking is slightly different from the Codalab results, as one of the teams retracted from the competition.} First, we note that all of the submissions outperform the three baselines for all the language pairs,\footnote{The only exception is HeyTUDa, which is outperformed by XMover-SHAP for De-Zh and by TransQuest-LIME for Et-En and Ro-En.} which indicates that error detection can indeed be approached as rationale extraction.

\paragraph{Approaches}
Overall, the submitted approaches vary a lot in the way they addressed the task. The following trends can be identified:
\begin{itemize}
    \item Following the recent standard in QE and similar multilingual NLP tasks, all the approaches rely on multilingual Transformer-based language models. 
    \item Submissions to the unconstrained track use the SOTA approach to word-level supervision explored previously by \citet{beringlab:wmt20}.
    \item The use of synthetic data produced by aligning MT outputs and reference translations from existing parallel corpora 
    proves an efficient strategy to identify translation errors. Supervising the predictions based on Transformer attention weights with the labels derived from synthetic data was used by the winning submission to the shared task.
    \item The approaches that rely on attention weights to predict human rationales (NICT-Kyoto and IST-Unbabel) achieve the best results for the constrained track.
    \item Both IST-Unbabel and HeyTUDa explore a wide set of explanation methods. The differences in performance are likely due to the method used for the final submission. While IST-Unbabel submission explores normalized attention weights, HeyTUDa use an ensemble of gradient-based approaches. A possible reason for the inferior performance of HeyTUDa is that the gradient is computed with respect to the embedding layer. As noted by \citet{fomicheva2021translation}, attribution to the embedding layer in the Transformer-based QE models does not provide strong results for the error detection task since word representations at the embedding layer do not capture contextual information, which is crucial for predicting translation quality.
    \item Gringham follow an entirely different strategy where they modify an existing reference-free metric to obtain both sentence score and word-level explanations in an unsupervised way. A similar approach is explored in our XMover-SHAP baseline, but the difference is that we apply SHAP explainer on top of XMover, while Gringham makes the XMoverScore inherently interpretable, which leads to better results. 
\end{itemize}  

\paragraph{Winners} 
The overall winner of the competition is the submission to the constrained track from NICT-Kyoto, which wins on 3 out of 4 language pairs, according to the source and target ranking. Fine-tuning on large amounts of synthetic data as well as the use of attention mechanism between the evaluation metric embeddings and the contextualized input representations seem to be the key to their performance. We note, however, that they offer a mixed approach with word-level supervision on synthetic data. Among the constrained approaches that do not use any supervision at word level, 
the best performing submission is IST-Unbabel, which outperforms other constrained submissions for all language pairs, except Ru-De, where they perform on par with Gringham on the target side and are surpassed by Gringham on the source side. For the unconstrained track we received only two submissions, from which IST-Unbabel* performs the best.

\paragraph{Sentence-level correlation} is not predictive of the performance of the submissions at detecting relevant tokens. This is due to the fact that submitted approaches vary in the role played by the sentence-level model. In fact, if we look at the submissions that follow comparable strategies, we do observe a correspondence between sentence-level and token-level results. For example, among the approaches that build upon a sentence-level QE model and use post-hoc methods to explain the predictions, IST-Unbabel tends to achieve higher performance both in terms of the token-level results and in terms of the Pearson correlation with sentence ratings, compared to HeyTUDa and the TransQuest-LIME baseline.


\paragraph{The performance on zero-shot language pairs} is lower than for Et-En and Ro-En. This is the case for all approaches except NICT-Kyoto on Ru-De, where the performance at word-level is comparable to the results for Et-En and Ro-En, even though the Pearson correlation for sentence scores is inferior. We attribute this outcome to the use of supervision with synthetic data, which helps boost performance for word-level QE when no manually labelled data is available, as has been shown by \citet{tuan2021quality}. Performance degradation for De-Zh is considerably larger than Ru-De. De-Zh was among the language pairs with the lowest inter-annotator agreement and, in addition, had a different distribution of sentence-level scores, with many high-quality translations, according to the annotators (see Section \ref{subsec:testset}).

\paragraph{Limitations of the evaluation settings}
Our current evaluation settings can be further improved in various ways. First, the submissions were ranked according to the global statistics, i.e.\ by comparing the mean AUC, AP and Rec-TopK scores of different submissions over a common set of test instances. However, such aggregation mechanisms ignore how many of its competitors a given submission outperforms and on how many test instances. In the future we plan to follow a more rigorous approach suggested by \citet{bt:2021} and use the Bradley–Terry (BT) model \cite{Bradley1952}, which leverages the instance-level pairing of metric scores. 


Second, the metrics used for evaluation are tailored for unsupervised explainability approaches that produce continuous scores, but they do not allow a direct comparison with the SOTA work on word-level QE, which is evaluated using F-score and Matthews correlation coefficient \cite{specia-etal-2020-findings-wmt}. One way to address this would be to require the participants to submit binary scores, but we discarded this option in this first edition of the shared task, as it would substantially limit the exploration of the explainability approaches. 

Finally, the binary rationales obtained from our pool of annotators through majority voting do not capture the fact that some words are more relevant for sentence-level quality than others. As shown in Table \ref{tab:example}, an alternative version of the data can be produced by averaging the scores assigned to each word by individual annotators, as an indication of the severity of translated errors. In the future, we plan to study to what extent such scores agree with the continuous explanation scores produced by the participants.
Another limitation of our annotation scheme is that sometimes a word may be missing in the machine translation, which can then not be highlighted (e.g., Russian does often not use determiners and the MT system may wrongly omit it when translating into English or German).

\section{Conclusions}

In this paper, we presented the findings of the Eval4NLP-2021 shared task on \emph{explainable Quality Estimation (QE)}, where the goal is to not only produce a sentence-level score for an MT output, given a source sentence, but also highlight erroneous words in the target (and source) sentence explaining the score. We detailed the data annotation, involving two novel non-English language pairs, our baselines (post-hoc explanation techniques on top of state-of-the-art QE models), as well as the participants' approaches to the task. These include supervised approaches, training on synthetic data as well as genuine post-hoc and inherent explainability techniques. 

The scope for future research is huge: for example, we aim to include new language pairs, especially low-resource ones, address explainability for metrics in other NLP tasks,
e.g.\ semantic textual similarity
\cite{agirre-etal-2016-semeval-2016}
and 
summarization \citep{gao-etal-2020-supert}, and 
identify error categories
of highlighted words, 
ideally in an unsupervised manner.

\section*{Acknowledgments}
Marina Fomicheva was supported by funding from the Bergamot project (EU H2020 Grant No. 825303). 
Piyawat Lertvittayakumjorn was supported by a scholarship from Anandamahidol Foundation.
We would like to thank Lisa Yankovskaya and Mark Fishel from the University of Tartu for helping organize and monitor the manual quality annotation. We also thank Anton Malinovskiy for adapting the Appraise interface for quality annotation with rationales. Finally, we gratefully thank the Artificial Intelligence Journal (\url{https://aij.ijcai.org/})
and Salesforce Research for their financial support enabling our human annotations. 

\bibliography{references}
\bibliographystyle{acl_natbib_nourl}



\end{document}